\PassOptionsToPackage{unicode}{hyperref}
\PassOptionsToPackage{hyphens}{url}
\documentclass[
]{article}
\usepackage{xcolor}
\usepackage{amsmath,amssymb}
\usepackage{iftex}
\ifPDFTeX
  \usepackage[T1]{fontenc}
  \usepackage[utf8]{inputenc}
  \usepackage{textcomp} % provide euro and other symbols
\else % if luatex or xetex
  \usepackage{unicode-math} % this also loads fontspec
  \defaultfontfeatures{Scale=MatchLowercase}
  \defaultfontfeatures[\rmfamily]{Ligatures=TeX,Scale=1}
\fi
\usepackage{lmodern}
\ifPDFTeX\else
\fi
\IfFileExists{upquote.sty}{\usepackage{upquote}}{}
\IfFileExists{microtype.sty}{% use microtype if available
  \usepackage[]{microtype}
  \UseMicrotypeSet[protrusion]{basicmath} % disable protrusion for tt fonts
}{}
\makeatletter
\@ifundefined{KOMAClassName}{% if non-KOMA class
  \IfFileExists{parskip.sty}{%
    \usepackage{parskip}
  }{% else
    \setlength{\parindent}{0pt}
    \setlength{\parskip}{6pt plus 2pt minus 1pt}}
}{% if KOMA class
  \KOMAoptions{parskip=half}}
\makeatother
\usepackage{color}
\usepackage{fancyvrb}

\DefineVerbatimEnvironment{Highlighting}{Verbatim}{commandchars=\\\{\}}
\newenvironment{Shaded}{}{}

\newcommand{\BuiltInTok}[1]{\textcolor[rgb]{0.00,0.50,0.00}{#1}}

\newcommand{\ControlFlowTok}[1]{\textcolor[rgb]{0.00,0.44,0.13}{\textbf{#1}}}

\newcommand{\DecValTok}[1]{\textcolor[rgb]{0.25,0.63,0.44}{#1}}

\newcommand{\FunctionTok}[1]{\textcolor[rgb]{0.02,0.16,0.49}{#1}}

\newcommand{\KeywordTok}[1]{\textcolor[rgb]{0.00,0.44,0.13}{\textbf{#1}}}
\newcommand{\NormalTok}[1]{#1}
\newcommand{\OperatorTok}[1]{\textcolor[rgb]{0.40,0.40,0.40}{#1}}

\newcommand{\StringTok}[1]{\textcolor[rgb]{0.25,0.44,0.63}{#1}}
\newcommand{\VariableTok}[1]{\textcolor[rgb]{0.10,0.09,0.49}{#1}}

\usepackage{longtable,booktabs,array}
\usepackage{calc} % for calculating minipage widths
\usepackage{etoolbox}
\makeatletter
\patchcmd\longtable{\par}{\if@noskipsec\mbox{}\fi\par}{}{}
\makeatother
\IfFileExists{footnotehyper.sty}{\usepackage{footnotehyper}}{\usepackage{footnote}}
\makesavenoteenv{longtable}
\NewDocumentCommand\citeproctext{}{}
\NewDocumentCommand\citeproc{mm}{%
  \begingroup\def\citeproctext{#2}\cite{#1}\endgroup}
\makeatletter
 \let\@cite@ofmt\@firstofone
 \def\@biblabel#1{}
 \def\@cite#1#2{{#1\if@tempswa , #2\fi}}
\makeatother
\newlength{\cslhangindent}
\newlength{\csllabelwidth}
\newenvironment{CSLReferences}[2] % #1 hanging-indent, #2 entry-spacing
 {\begin{list}{}{%
  \setlength{\itemindent}{0pt}
  \setlength{\leftmargin}{0pt}
  \setlength{\parsep}{0pt}
  \ifodd #1
   \setlength{\leftmargin}{\cslhangindent}
   \setlength{\itemindent}{-1\cslhangindent}
  \fi
  \setlength{\itemsep}{#2\baselineskip}}}
 {\end{list}}
\usepackage{calc}

\newcommand{\CSLLeftMargin}[1]{\parbox[t]{\csllabelwidth}{\strut#1\strut}}
\newcommand{\CSLRightInline}[1]{\parbox[t]{\linewidth - \csllabelwidth}{\strut#1\strut}}

\providecommand{\tightlist}{%
  \setlength{\itemsep}{0pt}\setlength{\parskip}{0pt}}
\usepackage{bookmark}
\IfFileExists{xurl.sty}{\usepackage{xurl}}{} % add URL line breaks if available
\hypersetup{
  pdftitle={The Responsibility Vacuum: Organizational Failure in Scaled Agent Systems},
  pdfauthor={Oleg Romanchuk; Roman Bondar},
  hidelinks,
  pdfcreator={LaTeX via pandoc}}

\title{The Responsibility Vacuum: Organizational Failure in Scaled Agent
Systems}
\author{Oleg Romanchuk \and Roman Bondar}
\date{}

\begin{document}
\maketitle

\subsection{Abstract}\label{abstract}

Modern CI/CD pipelines integrating agent-generated code exhibit a
structural failure in responsibility attribution. Decisions are executed
through formally correct approval processes, yet no entity possesses
both the authority to approve those decisions and the epistemic capacity
to meaningfully understand their basis.

We define this condition as \textbf{responsibility vacuum}: a state in
which decisions occur, but responsibility cannot be attributed because
authority and verification capacity do not coincide. We show that this
is not a process deviation or technical defect, but a structural
property of deployments where decision generation throughput exceeds
bounded human verification capacity.

We identify a \textbf{scaling limit} under standard deployment
assumptions, including parallel agent generation, CI-based validation,
and individualized human approval gates. Beyond a throughput threshold,
verification ceases to function as a decision criterion and is replaced
by ritualized approval based on proxy signals. Personalized
responsibility becomes structurally unattainable in this regime.

We further characterize a \textbf{CI amplification dynamic}, whereby
increasing automated validation coverage raises proxy signal density
without restoring human capacity. Under fixed time and attention
constraints, this accelerates cognitive offloading in the broad sense
\citeproc{ref-risko2016CognitiveOffloading}{{[}1{]}} and widens the gap
between formal approval and epistemic understanding. Additional
automation therefore amplifies, rather than mitigates, the
responsibility vacuum.

We conclude that unless organizations explicitly redesign decision
boundaries or reassign responsibility away from individual decisions
toward batch- or system-level ownership, responsibility vacuum remains
an invisible but persistent failure mode in scaled agent deployments.

\subsection{1. Introduction}\label{introduction}

Modern software deployments increasingly rely on agent-generated code
integrated through standard CI/CD pipelines. In a typical workflow,
autonomous agents generate changes, automated checks validate selected
properties, and human reviewers provide formal approval before
deployment. This pattern is widely adopted because it preserves familiar
governance structures while enabling higher development throughput.

At low decision volumes, this model functions as intended. Human
reviewers examine changes directly, engage with primary artifacts, and
their approvals reflect substantive understanding. As deployment volume
increases, however, the same approval mechanism is applied under
conditions where meaningful verification can no longer be sustained.

This paper analyzes what occurs when decision generation throughput
systematically exceeds human verification capacity.

\subsubsection{1.1 The Organizational
Problem}\label{the-organizational-problem}

We observe a recurring failure mode in scaled agent deployments:

\textbf{Formal authority persists while epistemic capacity does not.}

Human reviewers retain decision authority: their approval is required,
their identity is recorded, and organizational processes recognize them
as decision-makers. At the same time, the conditions required for
meaningful understanding --- time, attention, and access to primary
artifacts --- are no longer available at scale. Decisions continue to be
made, but no entity simultaneously possesses authority and capacity.

We refer to this condition as \textbf{responsibility vacuum}: a state in
which decisions occur, yet responsibility cannot be meaningfully
attributed because authority and verification capacity do not coincide.
The decision is formally approved, but not substantively owned.

The term ``responsibility vacuum'' has previously been used in empirical
studies of AI deployment in healthcare to describe fragmented
responsibility attribution in practice
\citeproc{ref-owens2025Managingresponsibilityvacuum}{{[}2{]}}. In this
work, we provide a structural and organizational characterization of
this phenomenon under scaling conditions.

This failure mode is distinct from:

\begin{itemize}
\tightlist
\item
  \textbf{Process failure}, as prescribed procedures are followed
  correctly;
\item
  \textbf{Technical defects}, as the underlying systems function as
  designed;
\item
  \textbf{Human error}, as no individual deviation is required.
\end{itemize}

Responsibility vacuum is not an exception to correct operation in such
systems. It is the outcome that correct operation produces under scaling
when decision authority remains individualized while verification
capacity is bounded.

\subsubsection{1.2 The Throughput-Capacity
Gap}\label{the-throughput-capacity-gap}

At the core of the problem lies a mismatch between the rate at which
decisions are produced and the rate at which they can be meaningfully
verified.

Let \(G\) denote the rate at which decisions requiring approval are
generated, and let \(H\) denote the rate at which a human actor can
meaningfully verify such decisions.

When \(G \le H\), reviewers can engage directly with decisions,
verification remains substantive, and responsibility can be attributed.
As \(G\) exceeds \(H\), the time and attention available per decision
shrink, and verification degrades. When \(G \gg H\), verification ceases
to function as a decision criterion: approval persists as a formal
requirement, increasingly grounded in proxy signals rather than
understanding.

This gap does not arise from poor practices or individual failure. It
reflects a basic asymmetry: human verification operates under bounded
time, attention, and cognitive bandwidth, whereas agent-driven decision
generation scales through parallelism and task decomposition. Once this
asymmetry becomes large enough, the character of review changes
qualitatively, setting the stage for the failure modes analyzed in the
sections that follow.

\subsubsection{1.3 Contributions}\label{contributions}

This paper makes the following contributions:

\begin{enumerate}
\def\labelenumi{\arabic{enumi}.}
\tightlist
\item
  We characterize \textbf{responsibility vacuum} as an
  authority--capacity mismatch and show that it is a structural property
  of scaled deployments, not a process deviation or component failure.
\item
  We identify a \textbf{scaling limit} under standard deployment
  assumptions, demonstrating that personalized responsibility cannot be
  preserved once decision throughput exceeds human verification
  capacity.
\item
  We describe a \textbf{CI amplification dynamic}, explaining why
  additional automated validation increases approval throughput while
  further disconnecting approval from understanding.
\item
  We outline \textbf{deployment implications}, arguing that explicit
  boundary redesign is required to make responsibility allocation
  visible and governable.
\end{enumerate}

\subsubsection{1.4 Scope}\label{scope}

This work presents an organizational analysis rather than a technical
audit. We do not claim that agent systems are poorly engineered, that
CI/CD pipelines are flawed, or that automation should be reduced. We do
not argue that human review must bottleneck all deployments.

We claim instead that the prevailing deployment paradigm encounters a
structural limit at scale, that this limit is organizational rather than
technical, and that addressing it requires redesigning responsibility
boundaries rather than optimizing individual components.

\subsubsection{1.5 Relationship to Prior
Work}\label{relationship-to-prior-work}

In prior work \citeproc{ref-romanchuk2026SemanticLaunderingAI}{{[}3{]}},
we analyzed \textbf{semantic laundering}, an architectural failure mode
in which propositions acquire unwarranted epistemic status through tool
boundary crossings. The present paper addresses a distinct but related
phenomenon: the organizational consequence of such epistemic gaps under
scaling.

Responsibility vacuum does not depend on any specific technical
mechanism. It arises whenever verification signals substitute for
understanding under bounded human capacity. Semantic laundering is one
mechanism by which this substitution occurs, but the analysis here
applies regardless of its source.

While semantic laundering concerns epistemic justification within agent
architectures, responsibility vacuum concerns organizational attribution
of decisions under scaling. The two results are orthogonal in scope.

\subsubsection{1.6 Methodological
Position}\label{methodological-position}

This paper combines formal characterization with organizational
analysis. While we introduce precise definitions and scaling arguments,
our objective is not theorem proving but structural diagnosis. The
constraints we analyze arise from bounded human capacity and unbounded
decision throughput and persist across tools, processes, and governance
models. Formalization is used to clarify these constraints, not to
reduce them to purely technical artifacts.

\subsection{2. Background}\label{background}

\subsubsection{2.1 Agent Deployment
Patterns}\label{agent-deployment-patterns}

Modern agent orchestration frameworks coordinate LLM-based code
generation workflows within standard software delivery pipelines. These
systems typically manage:

\begin{itemize}
\tightlist
\item
  task decomposition and execution sequencing,
\item
  state coordination across agent iterations,
\item
  completion detection via protocol-level markers,
\item
  aggregation of outputs for downstream validation and approval.
\end{itemize}

A critical property of such orchestrators is that they implement
\textbf{coordination contracts}, not \textbf{verification contracts}.
They ensure that an agent followed a prescribed interaction protocol and
reached a declared terminal state. They do not establish that the
produced output is correct, sufficient, or aligned with deployment
intent.

This distinction is intentional rather than defective. Coordination and
verification are categorically different responsibilities. Orchestrators
are designed to manage process flow, not to ground epistemic warrant.
Organizational failure arises when coordination completion is treated as
a substitute for verification completion.

\subsubsection{2.2 The CI Pipeline}\label{the-ci-pipeline}

Continuous Integration (CI) pipelines perform automated validation of
selected executable properties within a defined environment. Typical CI
checks include:

{\def\LTcaptype{none} % do not increment counter
\begin{longtable}[]{@{}ll@{}}
\toprule\noalign{}
CI validates & CI does not validate \\
\midrule\noalign{}
\endhead
\bottomrule\noalign{}
\endlastfoot
Syntax correctness & Semantic correctness \\
Existing tests pass & Test sufficiency \\
Build succeeds & Architectural soundness \\
Lint rules satisfied & Intent alignment \\
\end{longtable}
}

CI validates \textbf{what has been specified, where it has been
specified}, and \textbf{only under the conditions encoded in the
pipeline}. It does not validate whether the specified checks are
adequate, whether they cover relevant failure modes, or whether the
resulting system behavior is acceptable.

A successful CI run therefore establishes that predefined checks passed.
It does not establish that a change is correct, safe, or understood.

\subsubsection{2.3 Approval as Ritualized
Verification}\label{approval-as-ritualized-verification}

Prior to widespread CI adoption, human reviewers often executed
validation steps directly, such as running tests or inspecting runtime
behavior. Approval followed from first-hand engagement with primary
artifacts.

With CI integration, the formal approval action remains unchanged, while
its epistemic basis shifts. Reviewers increasingly rely on CI outcomes
as sufficient justification for approval. The decision action (approve
or reject) is preserved, but the underlying verification activity is
substituted by proxy confirmation.

This substitution is not reflected in audit records. Approvals grounded
in direct inspection and approvals grounded in proxy signals are
indistinguishable at the process level. The system records the same
approval event regardless of whether substantive understanding occurred.

As a result, verification transitions from an epistemic activity to a
ritualized procedural step, setting the stage for responsibility vacuum
under scaling.

\subsection{3. Model}\label{model}

\subsubsection{3.1 Core Terms}\label{core-terms}

\textbf{Authority}: Entity \(E\) has authority over decision \(D\) if
\(E\) holds formal rights to trigger \(D\) and \(D\) produces an
irreversible or externally binding system state, such that
organizational structures recognize \(E\) as the decision-maker,
independent of \(E\)'s epistemic understanding of \(D\).

\textbf{Capacity}: Entity \(E\) has capacity for decision \(D\) if \(E\)
can, within the available decision window, reconstruct a justified model
of \(D\) --- including its inputs, transformations, and plausible
failure modes --- to an extent sufficient to warrant responsibility for
\(D\).

Capacity is bounded by:

\begin{itemize}
\tightlist
\item
  \textbf{Temporal limits}: finite time available per decision
\item
  \textbf{Cognitive limits}: finite attention and working memory
\item
  \textbf{Epistemic access}: availability of primary artifacts (code,
  execution traces, domain context)
\end{itemize}

Capacity is not a function of expertise alone. It is a function of time,
access, and cognitive bandwidth under load.

\textbf{Responsibility vacuum}: A system exhibits a responsibility
vacuum for decision \(D\) if \(D\) occurred (i.e., an irreversible
action was taken) and no entity \(E\) exists such that \(E\) has both
authority over \(D\) and capacity for \(D\).

\[
\begin{aligned}
\text{ResponsibilityVacuum}(D) \;\Leftrightarrow\;
    &\text{Occurred}(D) \;\land \\
    &\forall E:\; \neg\bigl(\text{Authority}(E, D) \;\land\; \text{Capacity}(E, D)\bigr)
\end{aligned}
\]

The definition is existentially negative: responsibility vacuum is
defined by the absence of any entity satisfying both predicates, not by
the failure of a specific role.

\textbf{Ritual review}: Review in which the reviewer's action
(approve/reject) is decoupled from their understanding. The signature
persists; the understanding does not.

Ritual review emerges when approval actions are preserved as formal
requirements while the epistemic basis for those actions is
systematically substituted by proxy signals.

\subsubsection{3.2 Throughput Parameters}\label{throughput-parameters}

\textbf{Generation throughput (G)}: decisions requiring approval per
unit time.

\textbf{Verification capacity (H)}: decisions an entity can meaningfully
verify per unit time.

Generation throughput \(G\) is unbounded in principle, scaling with
agent parallelism and task decomposition. Verification capacity \(H\) is
bounded in principle, scaling at best linearly with human resources.

Meaningful verification requires the verifier to understand what they
are approving --- the change content, implications, and risk. Checking
``CI green'' without understanding the change is not meaningful
verification. Proxy confirmation (e.g., observing ``CI green'') without
engagement with primary artifacts does not constitute meaningful
verification.

This notion is intentionally normative rather than operationalized; the
analysis concerns responsibility attribution rather than the measurement
of verification quality.

\subsubsection{3.3 Scaling Limit}\label{scaling-limit}

Under standard deployment conditions, personalized responsibility
becomes structurally impossible when throughput exceeds capacity.

\textbf{Conditions}:

\begin{enumerate}
\def\labelenumi{\arabic{enumi}.}
\tightlist
\item
  Generation throughput \(G\) can exceed any fixed human capacity \(H\)
\item
  Approval requires formal human action (e.g., recorded sign-off such as
  a signature or an ``LGTM'' approval)
\item
  No batch-level ownership role exists (each decision requires
  individual approval)
\item
  Organizational pressure favors throughput over verification depth
\item
  Decision authority is attributed at the individual-decision level
  rather than the batch- or system-level.
\end{enumerate}

\textbf{Derivation}:

At \(G \le H\), the reviewer can understand each decision. Authority and
capacity coincide. Responsibility is attributable.

At \(G > \tau H\), the system crosses a qualitative threshold: time per
decision falls below the minimum required for epistemic reconstruction.
Beyond this point, verification cannot be partially preserved as a
decision criterion --- it ceases to function as such and is replaced by
ritualized proxies. This is a phase transition in the decision regime,
not a gradual loss of quality.

At \(G \gg H\), ritual review dominates. The reviewer retains authority
but lacks capacity. A responsibility vacuum emerges.

The threshold \(\tau\) is deployment-specific and depends on decision
complexity, reviewer expertise, and tooling. Its existence is
structural: for any fixed \(H\), there exists a rate \(G\) such that the
ratio \(G/H\) exceeds this threshold. The argument depends on the
existence of \(\tau\), not on its precise value. No empirical
calibration of \(\tau\) is required for the claim to hold.

\textbf{Scope}: This applies to decisions with irreversible or
high-cost-of-error effects (merge, deploy, payment). Reversible or
low-stakes actions are outside scope.

\subsubsection{3.4 Structural Invariance}\label{structural-invariance}

The responsibility vacuum is invariant under local optimizations.
Improvements in tooling, reviewer training, or signal quality ---
including the addition of automated verification signals --- may shift
the threshold \(\tau\), but cannot eliminate the existence of a regime
in which \(G \gg H\). As long as authority remains individualized and
capacity remains bounded, the vacuum re-emerges at scale.

\subsection{4. The CI Amplification
Dynamic}\label{the-ci-amplification-dynamic}

This section does not introduce a new failure mode. It analyzes a
concrete amplification mechanism through which the structural limits
identified in Section 3 manifest in modern CI/CD deployments.

The core claim is simple: CI does not resolve the authority--capacity
mismatch. It accelerates its realization by reshaping how verification
is performed under bounded capacity.

\subsubsection{4.1 The Intuition}\label{the-intuition}

``Add more CI checks to ensure quality.''

This is the standard organizational response to deployment failures. CI
pipelines do increase correctness guarantees by automating specific
forms of validation. However, in systems already operating near or
beyond human verification capacity, additional automation does not
restore responsibility.

Instead, it accelerates the transition to regimes in which approval
persists as a formal act while understanding ceases to function as a
decision criterion.

\subsubsection{4.2 Proxy Substitution Under Bounded
Capacity}\label{proxy-substitution-under-bounded-capacity}

Adding CI checks increases the density of automated validation signals
presented to the reviewer while leaving human verification capacity
unchanged:

\begin{verbatim}
More checks → More proxy signals → Less direct inspection
                                 → Less epistemic engagement
                                 → Wider responsibility gap
\end{verbatim}

Under fixed time and attention budgets, review effort shifts toward the
cheapest available signals compatible with maintaining throughput. Proxy
confirmations (e.g., ``CI green'') are strictly cheaper to consume than
primary artifacts such as code diffs, execution traces, or domain
reasoning. This reflects well-documented tendencies toward reliance on
automated cues under bounded cognitive resources
\citeproc{ref-parasuraman1997HumansAutomationUsea}{{[}4{]}}.

As proxy signal density increases without a corresponding increase in
verification capacity, engagement with primary artifacts is
systematically displaced. Primary inspection becomes residual rather
than central to the approval decision.

\textbf{Consequence}: Under fixed verification capacity, increasing
proxy density reduces the fraction of decisions reviewed against primary
artifacts.

This is not a claim about CI quality. CI correctly validates what it
validates. The effect concerns organizational behavior under bounded
capacity: when additional proxy signals are introduced without
additional time or attention, verification effort is reallocated away
from primary artifacts toward proxies.

\subsubsection{4.3 Capacity Compression via Epistemic
Substitution}\label{capacity-compression-via-epistemic-substitution}

The effect of CI amplification is not limited to reallocating
verification effort within a fixed capacity. It also alters the
effective verification capacity itself.

As defined in Section 3, verification capacity is not determined by time
alone. It depends critically on epistemic access to primary artifacts
--- the ability to reconstruct what a decision does, why it does so, and
how it can fail. When proxy signals become the dominant objects of
review, primary artifacts cease to function as routine inputs to
verification.

Over time, this substitution reshapes the verification regime. Review
processes, expectations, and norms adapt to proxy consumption. Access to
primary artifacts remains nominally available, but ceases to be
operationally central. As a result, the effective verification capacity
of reviewers decreases even if headcount and nominal time budgets remain
unchanged.

CI amplification therefore compresses capacity in two ways
simultaneously:

\begin{itemize}
\tightlist
\item
  by reallocating verification effort toward cheaper proxy signals;
\item
  by degrading epistemic access, which is a constitutive component of
  capacity itself.
\end{itemize}

This dynamic accelerates the transition into the regime described in
Section 3. Once generation throughput exceeds effective verification
capacity, authority remains attached to individual approvals while the
capacity required for responsibility attribution no longer exists. The
responsibility vacuum is not mitigated by CI. It is reached faster.

In the terms of Section 3, CI amplification effectively reduces
verification capacity \(H\) by displacing epistemic access, or
equivalently shifts the threshold \(\tau\) toward lower throughput
regimes.

\subsection{5. Case Study: Coordination-Only Agent
Orchestration}\label{case-study-coordination-only-agent-orchestration}

The following examples are schematic and intended for conceptual
illustration of coordination-only orchestration patterns rather than
direct correspondence to a specific production system.

\textbf{Note.} This section provides a concrete architectural
instantiation of the model introduced in Sections 3--4. The
responsibility vacuum follows from the stated assumptions and does not
require empirical validation. The examples below demonstrate how the
failure mode necessarily arises in common agent orchestration
architectures once decision throughput exceeds human verification
capacity.

\subsubsection{5.1 Coordination Contract vs Verification
Contract}\label{coordination-contract-vs-verification-contract}

Consider a typical agent orchestration runtime used to integrate
LLM-generated code into CI/CD pipelines. The orchestrator is responsible
for coordinating agent execution and determining when a task is
complete.

A minimal but architecturally representative completion contract is
shown below:

\begin{Shaded}
\begin{Highlighting}[]
\KeywordTok{class}\NormalTok{ Session:}
    \KeywordTok{def} \FunctionTok{\_\_init\_\_}\NormalTok{(}\VariableTok{self}\NormalTok{, output, open\_tasks, confirmations):}
        \VariableTok{self}\NormalTok{.output }\OperatorTok{=}\NormalTok{ output}
        \VariableTok{self}\NormalTok{.open\_tasks }\OperatorTok{=}\NormalTok{ open\_tasks}
        \VariableTok{self}\NormalTok{.confirmations }\OperatorTok{=}\NormalTok{ confirmations}

    \KeywordTok{def}\NormalTok{ has\_open\_tasks(}\VariableTok{self}\NormalTok{):}
        \ControlFlowTok{return} \BuiltInTok{len}\NormalTok{(}\VariableTok{self}\NormalTok{.open\_tasks) }\OperatorTok{\textgreater{}} \DecValTok{0}


\KeywordTok{def}\NormalTok{ is\_complete(session: Session) }\OperatorTok{{-}\textgreater{}} \BuiltInTok{bool}\NormalTok{:}
\NormalTok{    has\_marker }\OperatorTok{=} \StringTok{"COMPLETE"} \KeywordTok{in}\NormalTok{ session.output}
\NormalTok{    tasks\_done }\OperatorTok{=} \KeywordTok{not}\NormalTok{ session.has\_open\_tasks()}
\NormalTok{    stable }\OperatorTok{=}\NormalTok{ session.confirmations }\OperatorTok{\textgreater{}=} \DecValTok{2}

    \ControlFlowTok{return}\NormalTok{ has\_marker }\KeywordTok{and}\NormalTok{ tasks\_done }\KeywordTok{and}\NormalTok{ stable}
\end{Highlighting}
\end{Shaded}

This logic establishes \textbf{protocol completion}:

\begin{itemize}
\tightlist
\item
  the agent declared completion,
\item
  no pending tasks remain in the orchestration state,
\item
  the declaration is stable across iterations.
\end{itemize}

Crucially, this contract does not establish any epistemic warrant
regarding the produced output. It does not verify that tests were
executed, that code compiles, or that the implementation satisfies the
specification. The orchestrator fulfills its coordination responsibility
without contributing to verification capacity \(H\).

In the terminology of Section 3, the orchestrator produces a decision
candidate while leaving verification unchanged.

\subsubsection{5.2 Format-as-Verification as Structural
Pattern}\label{format-as-verification-as-structural-pattern}

Downstream components often treat agent-reported status fields as
verification signals. A minimal representation of this pattern is:

\begin{Shaded}
\begin{Highlighting}[]
\KeywordTok{def}\NormalTok{ validate\_output(payload: }\BuiltInTok{str}\NormalTok{) }\OperatorTok{{-}\textgreater{}} \BuiltInTok{bool}\NormalTok{:}
    \ControlFlowTok{return} \StringTok{"tests: pass"} \KeywordTok{in}\NormalTok{ payload}
\end{Highlighting}
\end{Shaded}

Here, the presence of a syntactic marker is treated as evidence that a
corresponding verification step occurred. This is \textbf{format
validation}, not \textbf{content verification}. The string may correctly
summarize an executed test run, or it may merely assert one.

This pattern is not a coding error. Any mechanism that upgrades
agent-generated claims into verification signals without independent
execution exhibits the same structural property, regardless of
implementation quality or tooling sophistication. This design pattern is
analyzed in detail in an engineering case study of production agent
orchestrators \citeproc{ref-romanchuk2026YourAIAgent}{{[}5{]}}.

The architectural decision is to treat \emph{reported status} as
\emph{verified status}. No new epistemic access is introduced.

\subsubsection{5.3 Capacity Substitution Under
Scaling}\label{capacity-substitution-under-scaling}

The consequences of these contracts depend on the throughput regime.

\textbf{Low-throughput regime} (\(G \le H\)). Human reviewers
independently reconstruct epistemic warrant by inspecting code and
executing tests. Orchestrator output and CI signals function as
proposals or summaries. Authority and capacity coincide. Responsibility
is attributable.

\textbf{High-throughput regime} (\(G \gg H\)). Human verification
capacity is exhausted. Orchestrator outputs and CI signals become the
sole basis for approval. The reviewer's role collapses to proxy
confirmation. Authority remains attached to the reviewer, but capacity
is absent by definition.

Given that the decision \(D\) has occurred, the system satisfies the
remaining condition of responsibility vacuum (Section 3.1):

\[
\forall E:\; \neg\bigl(\text{Authority}(E, D) \land \text{Capacity}(E, D)\bigr)
\]

No component is malfunctioning. The orchestrator satisfies its
coordination contract. CI validates specified checks. The failure
emerges from the interaction between bounded capacity and unbounded
decision generation.

\subsubsection{5.4 Architectural
Interpretation}\label{architectural-interpretation}

This case study demonstrates three structural properties:

\begin{enumerate}
\def\labelenumi{\arabic{enumi}.}
\tightlist
\item
  \textbf{Coordination without verification is sufficient to generate
  decisions.} Orchestrators need not be epistemic actors to drive
  deployments.
\item
  \textbf{Verification signals substitute for understanding under load.}
  CI and agent-reported status fields provide proxy signals that
  dominate approval decisions once capacity is exceeded.
\item
  \textbf{Responsibility vacuum emerges without error or deviation.} The
  system operates as designed. Responsibility disappears because no
  entity simultaneously satisfies authority and capacity.
\end{enumerate}

The failure is therefore not attributable to implementation defects,
poor model quality, or insufficient tooling. It is an organizational
consequence of deploying coordination-only agent systems under scaling
conditions that exceed human verification capacity.

\subsection{6. Responsibility Attribution
Breakdown}\label{responsibility-attribution-breakdown}

In systems exhibiting responsibility vacuum, post-incident analysis
follows a characteristic pattern: attribution proceeds through formally
correct components but does not terminate in an epistemic subject.

Approval is attributed to a reviewer. The reviewer points to CI. CI
points to passing checks. Checks point to agent-reported completion.
Orchestration validates protocol termination. At no point does the chain
reach an entity that both authorized the decision and possessed the
capacity to understand it. In the architectures described in Section 5,
this attribution chain maps directly onto concrete components. The
reviewer relies on CI outcomes, CI relies on agent-reported status
fields, and the orchestrator validates protocol completion without
contributing any epistemic warrant.

This is not a process failure or an error condition. Each component
operates within its specified contract. The breakdown emerges because
authority is preserved while verification capacity is exhausted.
Responsibility becomes structurally undefined once decision throughput
exceeds verification capacity.

\subsection{7. Deployment Implications}\label{deployment-implications}

The responsibility vacuum is not resolved by local improvements. It
forces an explicit organizational choice.

\subsubsection{7.1 What Does Not Resolve the
Vacuum}\label{what-does-not-resolve-the-vacuum}

Improving agent quality, adding verification logic inside the
orchestrator, training reviewers, or expanding CI coverage does not
restore responsibility attribution. These interventions may shift
thresholds or improve specific failure rates, but they do not alter the
underlying structural condition: authority remains individualized while
verification capacity remains bounded.

\subsubsection{7.2 Forced Organizational
Trade-offs}\label{forced-organizational-trade-offs}

Under scaled agent deployments, organizations face a limited set of
options:

\textbf{Option 1: Constrain throughput.}

Limit parallelism so that decision generation remains within human
verification capacity. Responsibility is preserved, but the scaling
advantage of automation is forfeited.

\textbf{Option 2: Reassign responsibility at aggregate levels.}

Introduce batch- or system-level ownership roles responsible for
outcomes rather than individual decisions. Responsibility is
re-personalized, but requires new organizational structures and
acceptance of aggregate risk.

\textbf{Option 3: Accept explicit system autonomy.}

Grant deployment authority to automated systems and treat resulting
behavior as an organizational liability. This aligns authority with the
system components that effectively determine outcomes, but requires
legal and governance frameworks that are largely undeveloped. This
option does not resolve responsibility vacuum at the level of individual
decisions. Instead, it formalizes the vacuum by abandoning
individualized responsibility and shifting accountability to the system
or organization as a whole.

There is no cost-free resolution. The prevailing deployment paradigm
defaults to responsibility vacuum because it avoids making these
trade-offs explicit.

\subsection{8. Related Work}\label{related-work}

This work intersects with several established research directions but is
not reducible to any of them. We briefly position responsibility vacuum
relative to adjacent lines of work and clarify its distinct scope.

\textbf{Semantic laundering}
\citeproc{ref-romanchuk2026SemanticLaunderingAI}{{[}3{]}} describes an
architectural failure mode in which propositions acquire unwarranted
epistemic status through tool boundary crossings. That analysis operates
at the level of epistemic justification inside agent runtimes. The
present work addresses a different level of abstraction: the
organizational consequence of such epistemic gaps under scaling.
Responsibility vacuum does not depend on any specific laundering
mechanism; it arises whenever verification signals substitute for
understanding under bounded human capacity. Semantic laundering is
therefore one sufficient mechanism, but not a necessary condition, for
responsibility vacuum.

\textbf{Automation complacency}
\citeproc{ref-parasuraman1997HumansAutomationUsea}{{[}4{]}} has been
studied as a behavioral tendency of human operators to over-trust
automated systems. In contrast, we show that under conditions where
decision throughput exceeds human verification capacity(\(G \gg H\)),
reliance on automated proxies is not a cognitive bias but a structural
necessity. Responsibility vacuum persists even in the absence of
complacency, inadequate training, or operator error.

\textbf{Scaled agent deployments in practice.} Recent industry reports
on autonomous coding agents document the rapid scaling of parallel agent
execution and task decomposition, reaching regimes where manual
verification becomes infeasible without structural changes to approval
and ownership models
\citeproc{ref-lin2026Scalinglongrunningautonomous}{{[}6{]}}. These
observations motivate the throughput--capacity assumptions used in this
paper but do not themselves analyze responsibility attribution.

\textbf{Agent orchestration architectures.} Open-source orchestration
systems such as Ralph Orchestrator exemplify coordination-centric agent
runtimes that manage task state, completion detection, and workflow
progression without providing epistemic guarantees about output
correctness
\citeproc{ref-obrien2026mikeyobrienralphorchestrator}{{[}7{]}}. Such
systems illustrate the separation between coordination contracts and
verification contracts assumed in our model.

\textbf{High-frequency trading regulation}
\citeproc{ref-2010FindingsRegardingMarket}{{[}8{]}} provides a
historical analogue in which automation exceeded feasible human
oversight while nominal responsibility remained human-assigned.
Post-incident regulation introduced circuit breakers and systemic
controls after failures occurred. This parallel illustrates that
responsibility vacuum is not unique to AI systems but emerges whenever
automated decision generation outpaces oversight capacity.

\textbf{AI governance and responsibility gaps.} Empirical studies in
safety-critical domains such as healthcare report fragmented or poorly
assigned responsibility for AI system monitoring and outcomes,
explicitly identifying responsibility gaps in practice
\citeproc{ref-owens2025Managingresponsibilityvacuum}{{[}2{]}}. These
findings align with our organizational analysis but do not provide a
structural explanation for why such gaps persist under scaling. These
issues have also been discussed at a high level in AI governance
research \citeproc{ref-dafoe2018AIgovernanceresearch}{{[}9{]}}.

\textbf{System-level effects of AI productivity gains.} Large-scale
empirical studies show that AI tools can significantly increase
individual productivity while simultaneously reducing collective
scrutiny or diversity of attention, producing systemic effects that are
not apparent at the component level
\citeproc{ref-hao2026Artificialintelligencetools}{{[}10{]}}. Such
effects are consistent with the CI amplification dynamic described in
Section 4.

Taken together, these works address epistemic correctness, human
behavior, industrial scaling, regulation, and governance. Responsibility
vacuum identifies a distinct organizational failure mode that arises at
their intersection when authority is preserved while verification
capacity is exhausted. The analysis applies broadly to any domain
combining high-throughput automated decision generation with
individualized human approval.

\subsection{9. Conclusion}\label{conclusion}

This work identifies \textbf{responsibility vacuum} as a structural
failure mode in scaled agent deployments. When decision generation
throughput exceeds bounded human verification capacity, authority and
capacity structurally diverge. Decisions continue to be executed through
formally correct processes, but no epistemic subject remains who both
authorizes and understands them.

The result is negative by design. Responsibility vacuum is not a
consequence of insufficient tooling, inadequate training, or immature
automation. It arises under correct operation of contemporary deployment
architectures once scaling assumptions are satisfied. Improvements in
model quality, orchestration logic, or CI coverage may shift thresholds
but cannot restore personalized responsibility.

The significance of this result is not prescriptive but diagnostic. It
constrains the space of admissible responses. Organizations cannot
``optimize away'' responsibility vacuum; they must explicitly choose how
responsibility is reassigned, whether by constraining throughput,
aggregating ownership, or accepting system-level autonomy with
corresponding liability.

Responsibility vacuum therefore marks a boundary of the current
deployment paradigm. Beyond this boundary, responsibility does not fail
accidentally --- it becomes structurally undefined.

\subsection*{References}\label{references}
\addcontentsline{toc}{subsection}{References}

\protect\phantomsection\label{refs}

\end{document}